\def\year{2022}\relax
\documentclass[letterpaper]{article} 
\usepackage{aaai21}  
\usepackage{times}  
\usepackage{helvet} 
\usepackage{courier}  
\usepackage[hyphens]{url}  
\usepackage{graphicx} 
\usepackage{natbib}  
\usepackage{caption} 
\title{News Category Dataset}
\author {
    Rishabh Misra \\
}

\affiliations{
    Twitter, Inc \\
    r1misra@eng.ucsd.edu
}

\begin{document}

\maketitle

\begin{abstract}
People rely on news to know what is happening around the world and inform their daily lives. In today’s world, when the proliferation of fake news is rampant, having a large-scale and high-quality source of authentic news articles with the published category information is valuable to learning authentic news’ Natural Language syntax and semantics. As part of this work, we present a \emph{News Category Dataset} that contains around 210k news headlines from the year 2012 to 2022 obtained from HuffPost, along with useful metadata to enable various NLP tasks. In this paper, we also produce some novel insights from the dataset and describe various existing and potential applications of our dataset.
\end{abstract}

\section{Introduction}
News is a part of communication that keeps us informed of the changing events, issues, and characters in the world outside. The dissemination of news aids in the education of the general public on current events in and across the globe. It has a significant impact on the formation of a person’s views, which in turn influence what they do for the benefit of society. The purpose of journalism is thus to provide citizens with news they need to make the best possible decisions about their lives, their communities, their societies, and their governments. Hence, journalistic trends (or news coverage) across the world can provide great insights into what society deems important at any period of time. 

Uncovering such trends can be a boon to policy makers and global leaders in addressing foremost issues general public cares about. For example, the Indian media’s coverage of the price hike that followed the 2008 global food crisis, pressurized the government into introducing targeted strategies and policies to insulate the poor and vulnerable from the price shock. Submitting to panic, the government devised various measures like food stock to be sold at subsidized rates along with a public distribution system. With its wide reach, the news media plays a crucial role in accelerating policy debates through increasing public awareness and initiating dialogues, thereby helping set the policy agenda.

In order to aid discovering various journalistic trends, in this work we introduce \emph{News Category Dataset}. The usefulness of this dataset comes from the metadata associated with each piece of news headline. Specifically, we supply the category information and publication date for each news article, which allows for running powerful analysis to extract variety of insights. Furthermore, this dataset with news category information can help researchers with various Natural Language tasks like Semantic Tagging, Named Entity Recognition (NER), Word Sense Disambiguation, etc.

\noindent The main contributions are summarized as follows:
\begin{itemize}
\item We introduce a high-quality \emph{News Category Dataset}, and describe how we collected and processed it.
\item We produce various interesting insights from our dataset to showcase how it can be used.
\item We describe wide adoption of this dataset in the community and highlight some of the interesting work people have produced using this dataset.
\item Lastly, we describe some of the open applications of this dataset.
\end{itemize}

\section{Dataset}
We present the construction of the dataset in this section by going over aspects such as the discovery of sources and building the dataset.

\subsection{Data Source Discovery}
In order to collect large-scale and high-quality source of authentic news articles, we started exploring all the news websites keeping following criteria in mind:
\begin{itemize}
    \item the source should maintain an archive of their previously published articles.
    \item the source should publish news in sufficient volume to enable collection of large-scale data.
    \item the source should include metadata around published news like category in which they were published and the date of publication.
    \item the source should have sufficient text about the published news that can be collected cleanly. Many websites publish ads in between news stories which makes collecting news specific text really challenging.
\end{itemize}

During our discovery phase, we observed that Huffington Post\footnote{https://www.huffpost.com/}, an American news aggregator website, meets all of the set forth criteria. Specifically, it maintains archive of thousands of articles published on its website since 2012 and contains sufficient metadata like news category information, date published, and author of the article. Furthermore, the archive also contains short description of the news article, which directly tends to our fourth criteria of providing high-quality text regarding the published news articles.

\subsection{News Category Dataset}
In this section, we present\footnote{Dataset at https://rishabhmisra.github.io/publications/} information available in our \emph{News Category Dataset}. The unique contribution of this dataset is the availability of fine-grained category information for large number of articles along with their high-quality text. The dataset we obtained has a total of 210,294 news headlines that subsumes news articles published between the time period of 2012 and 2022. For each record in our dataset, we have following attributes available:
\begin{itemize}
    \item $category$: category in which the article was published.
    \item $headline$: the headline of the news article.
    \item $authors$: list of authors who contributed to the article.
    \item $link$: link to the original news article. 
    \item $short\_description$: Abstract of the news article.
    \item $date$: publication date of the article.
\end{itemize}

The link to the article is also included to ease extracting more text in case it is needed. The dataset contains articles from 42 categories. Table \ref{table1} and \ref{table2} note the top-10 and bottom-10 categories respectively in the dataset based on number of corresponding articles.

\begin{table}[ht]
\centering

\begin{tabular}{c|c}
\textbf{News Category} & \textbf{Number of Articles} \\
\hline
Politics & 35,602  \\
Wellness & 17,945  \\
Entertainment & 17,362  \\
Travel & 9,900  \\
Style \& Beauty & 9,814 \\
Parenting & 8,791  \\
Healthy Living & 6,694  \\
Queer Voices & 6,347  \\
Food \& Drink & 6,340  \\
Business & 5,992  \\

\end{tabular}

\caption{Top 10 categories in the dataset}
\label{table1}
\end{table}

\begin{table}[t]
\centering

\begin{tabular}{c|c}
\textbf{News Category} & \textbf{Number of Articles} \\
\hline
Arts & 1,509 \\
Environment & 1,444 \\
Fifty & 1,401 \\
Good News & 1,398 \\
U.S. News & 1,377 \\
Arts \& Culture & 1,339 \\
College & 1,144 \\
Latino Voices & 1,130 \\
Culture \& Arts & 1,074 \\
Education & 1,014 \\

\end{tabular}

\caption{Bottom 10 categories in the dataset}
\label{table2}
\end{table}

\subsection{Data Curation Method}
We make use of open-source tools like BeautifulSoup, Selenium, and Chrome Driver to curate the dataset. For collecting data from Huffington Post, we use Huffington Post’s archive link \footnote{https://www.huffingtonpost.com/archive/} as the base. In all the articles presented, we extract their link, category, headline, abstract, authors, and date published using BeautifulSoup API. Once that is done on one page, we simulate a button click action using Selenium to go to the next page and repeat the process. Among all the news categories present on Huffington Post, there were a couple of categories that had a very low article count. In order to publish a high-quality dataset, we remove all the articles from categories that had less than 1000 articles. This is to ensure that any Machine Learning model trained on this data is not affected by the data skew. Since the headline text comes from a professional news website and has reasonably good quality (no misspellings, short hands, etc.), we did not do any additional pre-processing on it.

\section{Exploratory Data Analysis}

\begin{figure}[ht]
    \centering
    \includegraphics[width=0.45\textwidth]{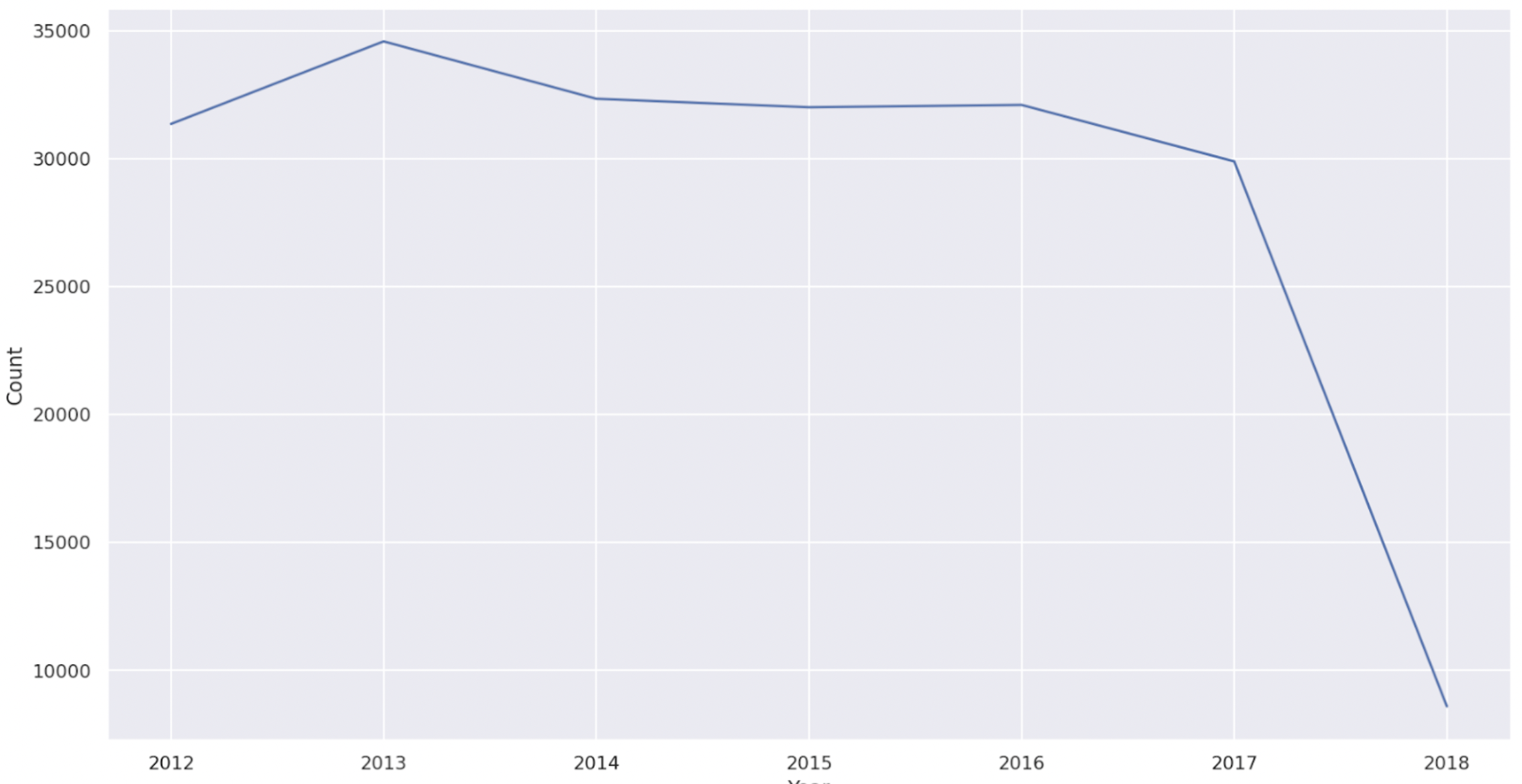}
    \caption{Number of articles published in each year between 2012 and 2018.}
    \label{fig1}
\end{figure}

As a basic exploration, we visualize the distribution of news articles from various years in figure \ref{fig1}. As a note, HuffPost stopped maintaining an extensive archive of the published news articles sometime after this dataset was first published in May 2018, so it is not possible to collect such a large scale dataset in the present day. Due to these changes in the website, there are about 200k headlines between 2012 and May 2018 and about 10k headlines between May 2018 and Sep 2022. Hence in figure \ref{fig1}, we see significantly fewer articles from 2018 onward and in case researchers are considering doing some temporal analysis, period of 2012 to 2017 is recommended.

\begin{figure}[ht]
    \centering
    \includegraphics[width=0.45\textwidth]{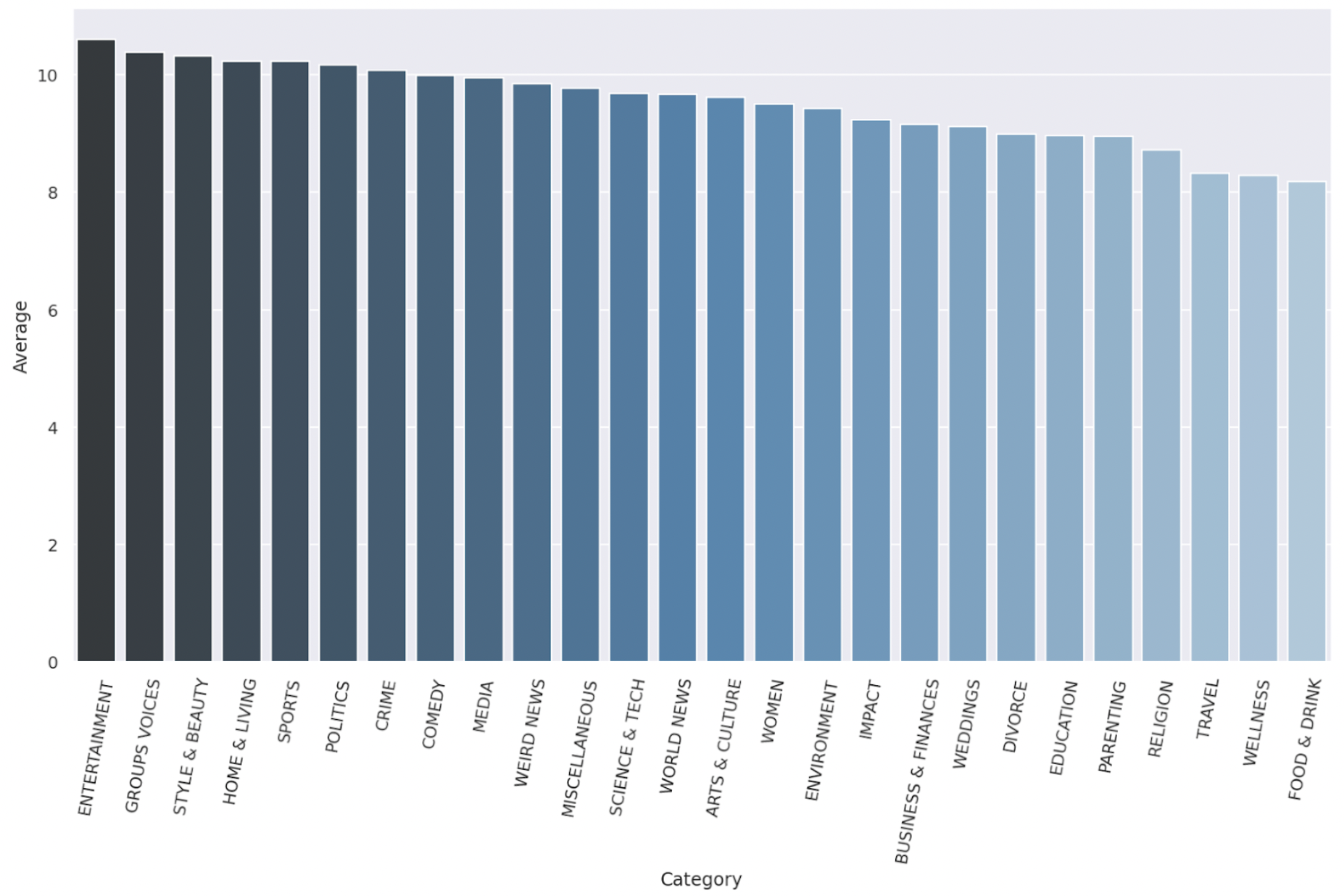}
    \caption{Average headline length for articles under specific categories.}
    \label{fig2}
\end{figure}

\begin{figure}[ht]
    \centering
    \includegraphics[width=0.45\textwidth]{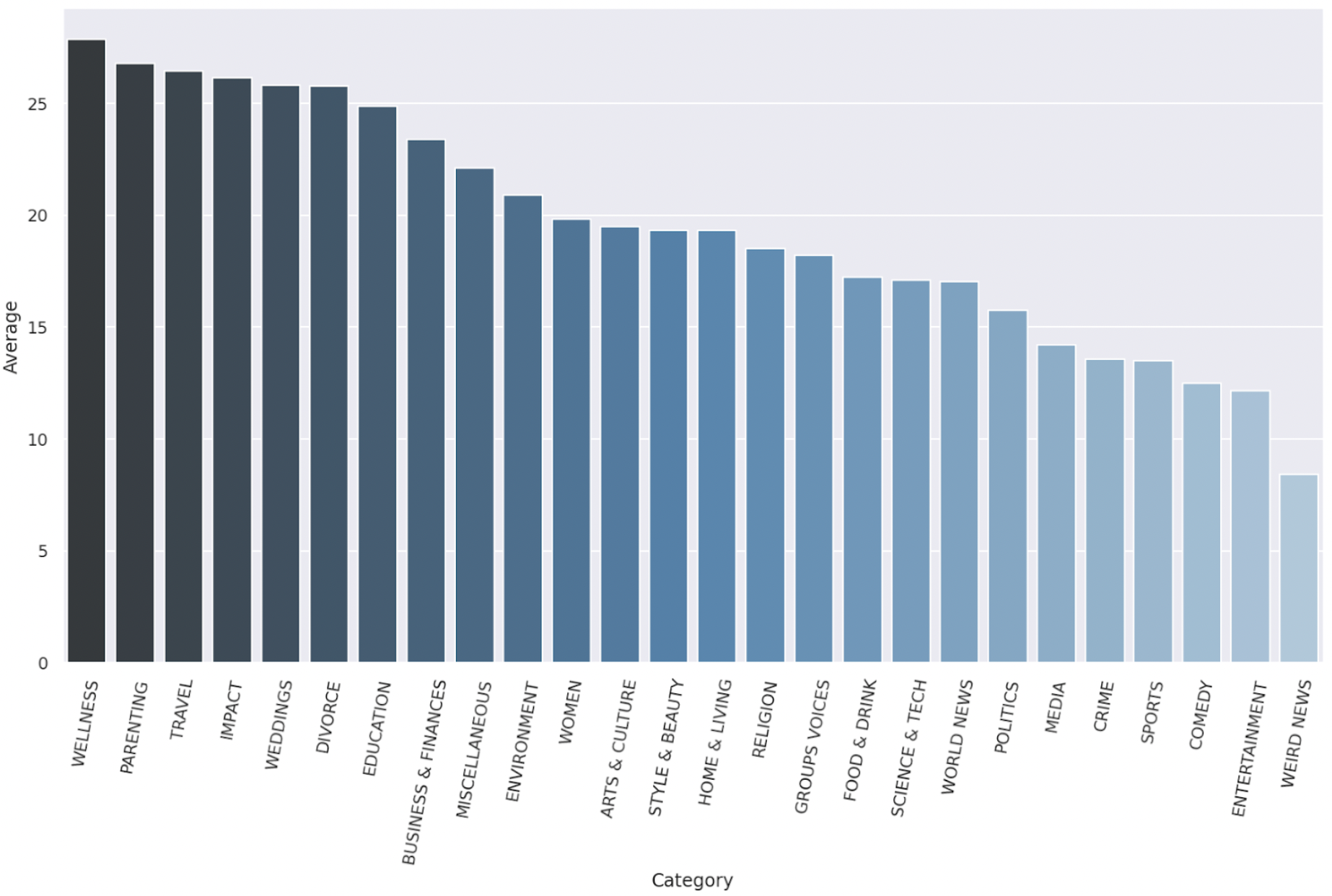}
    \caption{Average abstract length for articles under specific categories.}
    \label{fig3}
\end{figure}

In figures \ref{fig2} and \ref{fig3}, we showcase the average number of words present in the headline and short description of articles present in each category, respectively. Although differences in the headline length are not much, short description lengths have more variation.

\begin{figure}[ht]
    \centering
    \includegraphics[width=0.45\textwidth]{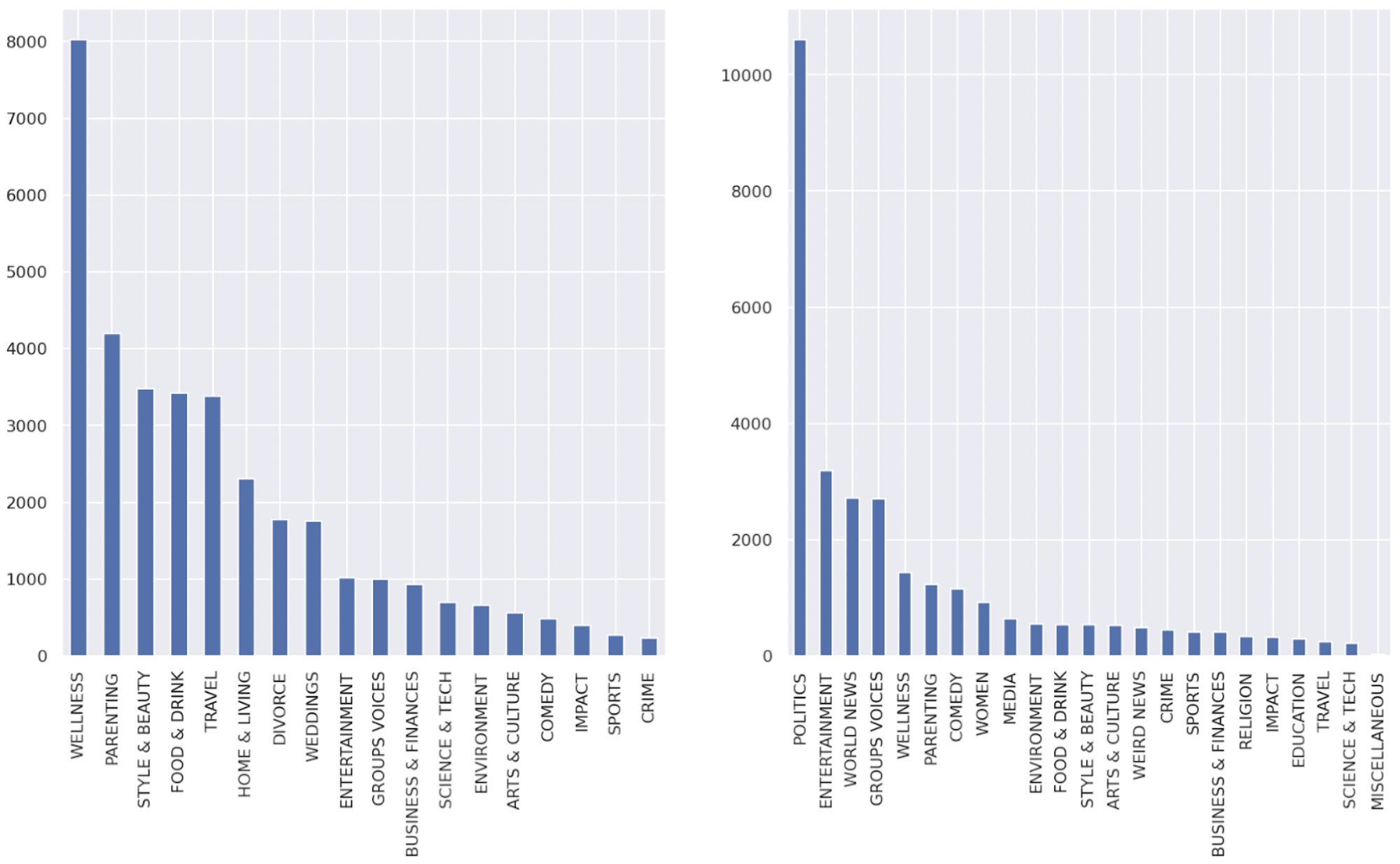}
    \caption{On the left is the distribution of news articles published in 2013. On the right is the distribution of news articles published in 2017.}
    \label{fig4}
\end{figure}

We also analyze the distribution of articles published under different categories for two different years 2013 and 2017 in figure \ref{fig4}. We see a fascinating insight: in 2013, the focus of reporting was a lot on Wellness, parenting, and beauty whereas, in 2017 the focus shifted to Politics, World News, and Entertainment. This clearly shows the shift in what's happening around the world over the years.

\begin{figure}[ht]
    \centering
    \includegraphics[width=0.45\textwidth]{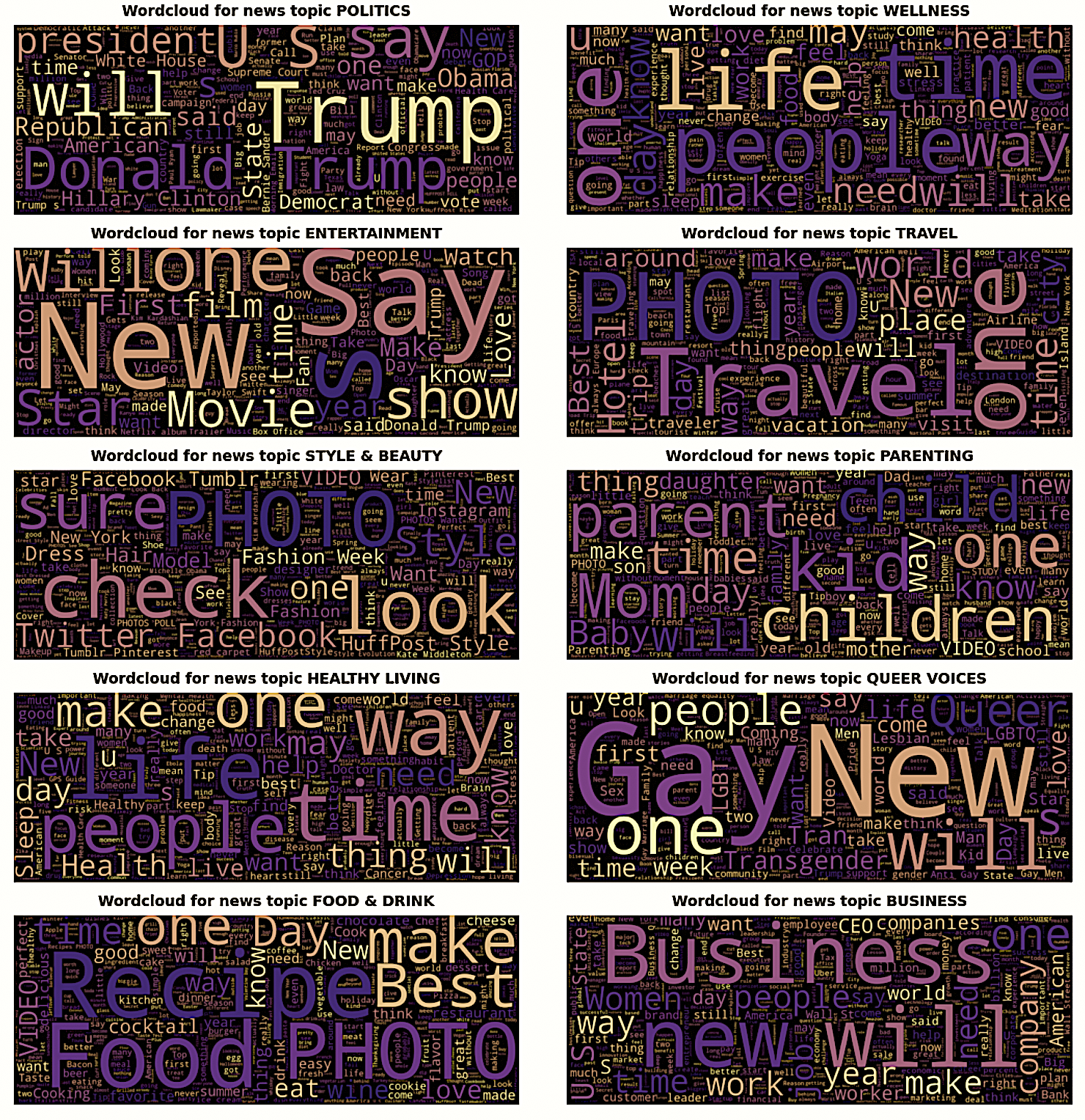}
    \caption{Word cloud of top-10 news categories in the dataset.}
    \label{fig5}
\end{figure}

Lastly, figure \ref{fig5} showcases word cloud of headlines from various news categories. We can visually see clear distinction in words used in each of the categories, which could be valuable to study various linguistic phenomenon.

\section{Applications}
The dataset was originally released in 2018 and it has seen wide adoption in the research community. In this section, we will cover some of the interesting applications to showcase how the dataset has been used so far to inspire further usage.

\subsection{Sarcasm Detection}
The earliest prominent use of this dataset comes from \cite{misra2018sarcasm, misra2019sarcasm}. Given News Category Dataset is a great source of authentic news, Misra et. al. perform a vertical integration of randomly sampled news headlines from News Category Dataset with news headlines collected from a sarcastic news website, TheOnion\footnote{https://www.theonion.com/}, to form a high-quality \emph{News Headlines Sarcasm Dataset}. Authors compare their dataset with other Twitter-based and hand labeled sarcasm datasets and note that News Headlines Sarcasm Dataset has less noise in terms of labels and language as compared to Twitter-based datasets and much more labels as compared to hand-labeled sarcasm datasets. They implemented a hybrid Neural Network architecture with attention-mechanism to extract various insights into what actually constitutes as sarcasm.

\subsection{Epidemiological Text Classification}
Another interesting use-case comes from \cite{mutuvi2020multilingual} who treat news articles contained in News Category Dataset as a great source of non-epidemiology articles by filtering out articles with direct or indirect mentions of disease outbreaks (e.g., plague, cholera, cough). They also perform vertical integration of this filtered data with epidemiology related news to form a new dataset for Epidemiological Text Classification. Authors ultimately explore various deep learning architectures to tackle this interesting linguistic problem.

\subsection{Gender Representation in News}
\cite{dacon2021does} use News Category Dataset to study bias issues in
gender representation in English news headlines. Specifically, authors
aim to detect and examine the phenomenon of implicit and explicit gender bias in the abstracts of news articles to gain a sense of understanding of the gender representation in the news by examining the relationships between social hierarchies and news content. They examine the news corpora for gender biases in distribution, content, and labeling and word choice, and demonstrate that there exist conclusive socially-constructed biases in regards to gender by introducing a series of measurements.

\subsection{The Media Language of Police Killings}
Researchers from National Bureau of Economic Research use News Category Dataset in \cite{moreno2022officer} to identify media language around police killings. They employ a BERT based model on News Category Dataset to identify the stories that were related to Crime category. Analysis done by authors on various news articles showcases that the news media is significantly more likely to use several language structures - e.g., passive voice, nominalization, intransitive verbs - that obfuscate responsibility for police killings compared to civilian homicides. Furthermore, their paper highlights the importance of incorporating the semantic structure of language, in addition to the amount and slant of coverage, in analyses of how the media shapes perceptions.

\subsection{Other Applications}
As we can see from previous section that our \emph{News Category Dataset} can be used in variety of ways to study important problems. Researchers can treat this as a generic source of authentic news or as a high-quality source of fine-grained category specific news. Overall, this corpus can act as a benchmark for Natural Language Processing tasks that involve understanding the syntax and semantics of different categories of text. Some potential open-ended use case for this dataset could be:
\begin{itemize}
    \item Tagging of untracked news articles.
    \item Analyzing how the focus of jornalism has shifted over the years.
    \item Analyzing articles across different categories in terms of writing style, presentation, etc.
    \item Tagging specific parts of the text with their tone corresponding to specific categories.
\end{itemize}
 
Kaggle community has also extensively worked with this dataset and some of the implementations can be explored here\footnote{https://www.kaggle.com/datasets/rmisra/news-category-dataset/code}.

\section{Concluding Remarks}
In this paper, we introduce a large-scale and high-quality \emph{News Category Dataset} that can be used as a benchmark for variety of Natural Language Processing tasks that involve understanding the syntax and semantics of general purpose news or news from specific categories. We also presented some novel insights from the dataset and describe various existing and potential applications of our dataset.

\bibliography{main}

\begin{thebibliography}{5}
\providecommand{\natexlab}[1]{#1}
\providecommand{\url}[1]{\texttt{#1}}
\providecommand{\urlprefix}{URL }
\expandafter\ifx\csname urlstyle\endcsname\relax
  \providecommand{\doi}[1]{doi:\discretionary{}{}{}#1}\else
  \providecommand{\doi}{doi:\discretionary{}{}{}\begingroup
  \urlstyle{rm}\Url}\fi

\bibitem[{Dacon and Liu(2021)}]{dacon2021does}
Dacon, J.; and Liu, H. 2021.
\newblock Does gender matter in the news? detecting and examining gender bias
  in news articles.
\newblock In \emph{Companion Proceedings of the Web Conference 2021}, 385--392.

\bibitem[{Misra(2018)}]{misra2018sarcasm}
Misra, R. 2018.
\newblock News Headlines Dataset For Sarcasm Detection.
\newblock \doi{10.13140/RG.2.2.16182.40004}.

\bibitem[{Misra and Arora(2019)}]{misra2019sarcasm}
Misra, R.; and Arora, P. 2019.
\newblock Sarcasm Detection using Hybrid Neural Network.
\newblock \emph{arXiv preprint arXiv:1908.07414} .

\bibitem[{Moreno-Medina et~al.(2022)Moreno-Medina, Ouss, Bayer, and
  Ba}]{moreno2022officer}
Moreno-Medina, J.; Ouss, A.; Bayer, P.; and Ba, B.~A. 2022.
\newblock Officer-Involved: The Media Language of Police Killings.
\newblock Technical report, National Bureau of Economic Research.

\bibitem[{Mutuvi et~al.(2020)Mutuvi, Boros, Doucet, Lejeune, Jatowt, and
  Odeo}]{mutuvi2020multilingual}
Mutuvi, S.; Boros, E.; Doucet, A.; Lejeune, G.; Jatowt, A.; and Odeo, M. 2020.
\newblock Multilingual epidemiological text classification: a comparative
  study.
\newblock In \emph{COLING, International Conference on Computational
  Linguistics}, 6172--6183.

\end{thebibliography}

\end{document}